\documentclass[11pt]{article}

\usepackage{epsfig}
\usepackage{a4}
\usepackage{nobi}
\usepackage{ims}
\usepackage{nobibib}
\usepackage{nobiDG}
\usepackage{html}

\newcounter{defcnt}
\newenvironment{definition}[2][]
        {\refstepcounter{defcnt}
                \par\addvspace{1em}
                \noindent
                {\bf Definition \thedefcnt} (#2):%
                \itshape%
	}
        {\par\addvspace{1em}%
	}

\begin{document}

\begin{titlepage}
\centering

{\large \bf A Projection Architecture for Dependency Grammar \\
	and How it Compares to LFG}

\vspace{0.5cm}
\nobi 

\vspace{0.5cm}
\nobijoborga \\ \nobijobstrasse \\ \nobijobplz\ \nobijobstadt \\
\textsc{\nobijobemail}

\vspace{2cm}
{\large \bf Proceedings of the LFG98 Conference} \\
\vspace{0.5cm}
The University of Queensland, Brisbane \\
\vspace{0.5cm}
Miriam Butt and Tracy Holloway King (Editors) \\

\vspace{2cm}
1998 \\
\vspace{0.5cm}
CSLI Publications \\
\vspace{0.5cm}
\textsc{http://www-csli.stanford.edu/publications}

\end{titlepage}

\begin{abstract}
This paper explores commonalities and differences between \dachs, a variant
of Dependency Grammar, and Lexical-Functional Grammar. \dachs\ is based on
traditional linguistic insights, but on modern mathematical tools, aiming
to integrate different knowledge systems (from syntax and semantics) via
their coupling to an abstract syntactic primitive, the dependency relation.
These knowledge systems correspond rather closely to projections in LFG. We
will investigate commonalities arising from the usage of the projection
approach in both theories, and point out differences due to the
incompatible linguistic premises. The main difference to LFG lies in the
motivation and status of the dimensions, and the information coded there.
We will argue that LFG confounds different information in one projection,
preventing it to achieve a good separation of alternatives and calling the
motivation of the projection into question. \\
\begin{rawhtml}
A PostScript version of this paper can be retrieved
<a href="http://www.ims.uni-stuttgart.de/~nobi/papers/lfg98/paper.ps">here</a>.
\end{rawhtml}
\end{abstract}

\section{Introduction
	\label{ch:intro}}

Dependency Grammar (DG) was introduced into modern linguistics by
\cite{Tesniere1959}. Since then, a number of quite different architectures
have been proposed based on the dependency relation \cite{Hudson1993}.
There are, e.g., rule-based vs. lexicalized approaches, non-derivational
vs. stratificational approaches with various levels, and proposals with
varying degrees of procedural specifications. Unfortunately from a
theoretical as well as computational point of view, all of them exhibit
some empirical and/or formal deficits.

On the other hand, the notion of valence and dependency has been
incorporated into all major syntactic theories based on phrase structure,
under the names of functional coherence and completeness, $\theta$-grid
\cite{Haegeman1994}, subcategorization list \cite{PS-II}, extended domain
of locality \cite{Joshi1995}, etc. These theories have in recent years
developed a strong formal base, often based on model theory; and
comparisons among them as well as implementations of parsers have
benefitted from this mathematical work. In addition, the trend to
lexicalization has resulted in a number of approaches replacing complex
rules systems by few combinatory operations, solely relying on lexical
information to express combinatory restrictions.

One aim of this paper, then, is to compare some ideas of LFG to an approach
based on the dependency relation as syntactic primitive, working out
commonalities (mainly found on the formal side), and identifying
differences. We will propose a specific theoretic architecture for DG which
postulates an abstract syntactic primitive, the dependency relation, and
which conceives other properties such as morphosyntax, ordering, and
semantics as consequences or, analytically, indicators of this dependency
relation. These premises will result in a completely lexicalized
architecture which allows to factor ambiguities from the syntactic
representation. To make things more precise, we will give a
state-of-the-art formal framework for DG, based on (multi-)modal logic.

The paper is structured as follows: First, we will briefly review current
DG practice, and take a look into model-theoretic approaches to syntax.
Then, we make some short remarks about LFG (Section~\ref{ch:LFG}) and
German (Section~\ref{ch:german}), which motivated this work. In
Sec.~\ref{ch:dachs} we to introduce \dachs\ (Dependency Approach to
Coupling Heterogeneous knowledge Sources), discussing the descriptional
dimensions and their linking. Finally, a number of parallels and
differences to LFG will be investigated in Sec.~\ref{ch:compare}. Our
conclusion will be that the projection architecture of LFG does not fully
exploit the advantages of the projection idea.

\subsection{Dependency Theory
	\label{ch:DG}}

A very brief characterization of DG is that it recognizes only lexical, not
phrasal nodes, which are linked by directed, typed, binary relations to
form a dependency tree \cite{Tesniere1959,Hudson1993}. Several formulations
assume a rule base determining dominance and precedence, e.g. \emph{Slot
Grammar} \cite{McCord1990}, while most of the DG variants lexicalize at
least the dominance information in valency frames. A number of them are
stratified, i.e., assume several representations linked by correspondence
rules. The most prominent in this class is surely
\emph{Meaning-Text-Theory} \cite{Melcuk1988}, which assumes seven strata of
representation ranging from semantic representation through unordered
dependency trees to morpheme sequences. Some of the rule types have not yet
been specified \cite[, p.187f]{Melcuk+Pertsov1987}. Another stratified DG
is \emph{Functional Generative Description} \cite{Sgall1986}, which assumes
a semantic level and an underlying syntactic representation
\cite{Petkevic1995}, which is mapped via ordering rules \cite{Kruijff1997}
to surface representation. Lexicalized, non-derivational accounts such as
\emph{Dependency Unification Grammar} or \emph{Lexicase} are very similar
to modern phrase-based theories. Dependency Unification Grammar
\cite{Hellwig1993} tries to provide dependency trees suitable for a
psychologically inspired inference model \cite{Hellwig1980}, but is based
on an operational semantics of its data types only. Lexicase
\cite{Starosta1988} employs \xbar-inspired dependency trees and formally
very simple word descriptions, namely fully specified feature sets. This
has the disadvantage, however, that many lexical ambiguities are required
to capture the many different environments a word may occur in. Word
Grammar \cite{Hudson1990} comes closest to our goals, being not only
lexicalized, non-derivational, and with syntactic and semantic
substructures, but additionally defining a propositional description
language. There are, however, formal inconsistencies in its word order
description and the inheritance mechanism as described in
\ct{Hudson1990} \cite{nobi:acl97}.

\subsection{Model-Theoretic Syntax
	\label{ch:model}}

Logical approaches to syntax may take the form of proof theories or model
theories \cite{Rogers+Cornell1997}. Proof theories view a grammar as a set
of axioms and deduction rules, and the question of grammaticality is
equivalent to the question of provability within the calculus. Categorial
Grammars are a prominent example in this class. On the other hand, model
theory emphasizes a priori structures, so-called models, which are
described by logical formulae. Here, the question of grammaticality is the
question whether a model exists which satisfies the descriptions of all
words of the utterance. Model-theoretic approaches may try to define new
models (e.g., logically equivalent, but simpler to process) for existing
theories \cite{Stabler1992}, or they may try to formalize models as they
already exist in the linguistic literature
\cite{Kasper+Rounds1990,Carpenter1992,Blackburn1993,Kracht1995,Rogers1996}.

Our approach falls into the last category. Following \cite{Blackburn-SLT},
we will use Kripke models to represent syntactic structures, and define a
multi-modal logic \cite{Fitting1984} for describing them. Basing the
formalization on modal logic has several welcome consequences. First, the
distinction between (logical) meta language and (graphical) object language
allows to compare the expressivity with other frameworks, such as
unification grammars \cite{Blackburn-SLT,Kracht1995}. For example,
non-functional modalities correspond to set-valued features in unification
approaches, both formally and in actual usage (e.g., for adjuncts). Second,
the formal precision allows to derive mathematical results on computational
complexity of recognition and generative capacity \cite{nobi:acl97}, which
contrasts previous results on rather impoverished DG conceptions
\cite{Gaifman1965,Lombardo1996}. Third, the linguistic descriptions in
terms of, e.g., word class or dependency are directly translatable into
modal propositions or operators, respectively.

\section{LFG and German Word Order}

\subsection{The LFG Architecture}
	\label{ch:LFG}

Given the audience of this talk, and my knowledge of LFG, I will not
attempt a summary or overview of LFG theory. I will rather point out which
properties of LFG triggered the comparison presented here. 

Some profound differences between LFG and \dachs\ make them unlikely
condidates for comparison: LFG employs a (usually large) explicit rule
base, while \dachs\ is completely lexicalized. LFG uses phrasal categories
in these rules, which have no status at all in \dachs. On the other hand,
there are several shared assumptions, such as the idea that well-formedness
corresponds to satisfiability of descriptions, and the surface-orientation
that eliminates underlying strata and derivations. 

The most prominent commonality, however, is the projection idea: LFG
defines a number of levels of representation which are formally different
and specialized to different types of information. These levels are linked
via structural correspondences, which map elements of one level to elements
of another level. The correspondence is not defined explicitly but rather
emerges from the joint statement of restrictions on these levels in rules
and lexical entries \cite{Kaplan1995}.

There have been different proposals as to the number, the content, and the
linking of these levels in LFG (compare \ct{Halvorsen1995,Butt1996}), and
here lies the difference to \dachs\ that will concern us most.

\subsection{German Word Order}
	\label{ch:german}

Again, given the audience, I will only remind you of very general
properties of word order in German, and take note of treatments in LFG, as
far as I have observed them.

Traditional descriptions of German word order make use of the notion of a
\key{positional field}, i.e., a topological position defined with reference
to the finite and infinite verb parts (which constitute the so-called
\key{Satzklammer}).

One confirmation for the concept of a topological field (contrasted with
the notion of constituent) comes from the fact that the topological field
does not have a categorial implication. That is to say, whereas a
constituent always is of a certain category, a topological field is (more
or less) neutral to the category of its element(s). For example, one finds
NPs, APs, VPs, PPs, $\overline{\mbox{S}}$s, and even separable verb
prefixes in the German Vorfeld.%
\footnote{%
As always, there are some exceptions to this rule: Relative clauses, which
may appear on their own in the Nachfeld, do not (on their own) occur in the
Vorfeld. To take another example, pronouns do not occur in the Nachfeld. 
}
Independent of any particular explanation of topicalization or
extraposition, it seems that the major category does not play an important
role in it.  

Now the notion of a topological field cannot be easily defined within LFG,
given its architecture. Unlike \xbar-theory, which defines categorially
unspecified positions, the context-free backbone of LFG requires one to
explicitly specify the categories allowed. This requires a huge
number of such rules which list all the different possibilities.%
\footnote{%
There are, of course, abbreviatory devices in any implementation of LFG
which allow a more concise presentation of the rules, but which to my
knowledge have no theoretical status in LFG.
}
Other means independent of the context-free rules are then used to restrict
the selection among the rules, most notably the conditions of functional
completeness and functional coherence. 

The first example of this reliance on restrictions on f-structure that came
to my attention is \cite{Netter1986}, who proposes for the VP a rule with a
maximal set of complements, each being optional (the proposal did not take
alternative orderings within Mittelfeld into account). Other proposals
amount to stating rules of the form \texttt{\mbox{S --> XP NP}}, where
\texttt{XP} is expanded into (nearly) all categories. Completeness and
coherence will ultimately weed out the invalid analyses, but the category
in these rules surely is part of the problem, not the solution.

Moving order constraints from c-structure to f-structure, as is proposed by
\ct{Zaenen+Kaplan1995}, does not solve the problem either.
\ct[:231]{Zaenen+Kaplan1995} acknowledge that c-structure in this setting
becomes underdetermined in the sense that there are no clear criteria for
distinguishing between alternative analyses on c-structure. We will see
below how one could define one level exclusively concerned with ordering
facts, such that the level is well-defined and the theory becomes more
modular.

\section{\dachs}
	\label{ch:dachs}

This section will present the basic design considerations underlying
\dachs, and then move to a description of the individual dimensions of
description.

\dachs\ developed in a project in text understanding and knowledge
extraction \cite{nobi:ijhcs94}. In this setting, processing efficiency and
incremental conceptual interpretation were of paramount importance; and we
did not relegate them (solely) to the parsing strategy, but took
precautions already in the grammar design. Both issues are addressed by the
idea of coupling knowledge systems: Lexical ambiguities can be reduced by
coupling several specialized knowledge systems, resulting in a smaller
search space for parsing, while incremental conceptual interpretation is
achieved through the coupling of syntax to a conceptual representation.

Summarizing these requirements and the discussion in Sec.~\ref{ch:DG}, we
require the following of our dependency grammar:

\begin{itemize}
\item to retain the traditional semantic motivation of dependencies (this
makes them less arbitrary and facilitates the conceptual interpretation of
syntactic structure),

\item to be surface-oriented and non-derivational (this avoids rather
arbitrary abstract underlying strata and facilitates incremental
processing),

\item to be strictly lexicalized (this eases grammar specification and may
make processing more efficient),

\item to define dimensions of descriptions which allow isolating
alternatives within the dimensions (this reduces lexical ambiguites and
improves processing efficiency),

\item not to assign priorities to individual dimensions as in many
stratificational approaches (again, this enhances incrementality),

\item to be declarative and formally precise (this is the basis for
theoretical investigations into the grammar's properties, for comparison
with other frameworks, and for computer implementations).
\end{itemize}

To fulfill these requirements, \dachs\ postulates an abstract syntactic
primitive, the dependency relation, which is linked to different
descriptional dimensions, such as morphosyntax, order, and semantics. We
conceive of them as consequences (or, analytically: indicators) of the basic
dependency relation. As we will argue, the information represented in each
dimension should be restricted to one type; otherwise, the dimension will
require complex and redundant specifications.

The following sections informally introduce the dependency tree as the
central dimension, and the word class and the word order domain as
dimensions mapped off the dependency tree. A more precise account based on
model theory is given in \cite{nobi:coling98}, where a modal logic is used
to describe the dependency structures. We will also ignore here the feature
annotations and the conceptual interpretation of the syntax tree; see
\cite{nobi:coling98,nobi:diss} for a complete account.

\subsection{Dependency Tree
	\label{ch:deptree}}

The dependency tree is the backbone of the syntactic representation. As
introduced in Sec.~\ref{ch:DG}, it consists of a set of word nodes, linked
by typed, binary, directed relations. The dependency relations together
form a rooted tree over the set of words. We do not require it to be
projective, because we assume semantically motivated dependencies, and word
order (e.g., various topicalization possibilities) will not allow these to
be projective. The dependency tree for the example sentence \os{Den Mann
hat der Junge gesehen} (\os{the man$_{ACC}$ -- has -- the boy$_{NOM}$ --
seen}) is shown in Fig.~\ref{fig:deptree}.

\begin{figure}
\centering
\epsfig{file=deptree.eps,width=0.4\columnwidth}
\caption{Example Dependency Tree 
	\label{fig:deptree}}
\end{figure}

Since we are not concerned with specific analyses here, but rather with the
formal architecture, we will not be specific about the dependency relation
types, but rather assume a set $\D = \{subj, obj, vpart, \ldots\}$ of
linguistically motivated types. A dependency relation of type $d \in \D$
will be written as $R_d$, and we will abbreviate the union $\bigcup_{d \in
\D} R_d$ as $R_\D$. If \W\ is the set of words, totally ordered by the
precedence relation \pre, we define a dependency tree as follows.

\begin{definition}{Dependency Tree (preliminary)}
A \emph{dependency tree} is a tuple \tuple{\W, w_r, R_\D} where
$R_\D$ forms a tree over $\W$ rooted in $w_r$.
\end{definition}

\subsection{Word Class}

A rather trivial descriptional dimension is the word class: Each word
(within some syntactic analysis) belongs to exactly one word class, which
for our present purposes is atomic, i.e., unanalyzable. We will encode word
class assignment by a function $V_\C$ mapping words to word classes, but it
should be clear that this mapping could also be presented as a `real'
projection, consisting of a set of objects (the word classes) onto which
the respective words are mapped. The point to note, however, is that
categorial restrictions are a priori independent of any other restriction:
One might require a certain category of a subordinated word without
requiring anything else of it. We call the set of word classes \C, and
define: 

\begin{definition}{Dependency Tree}
A \emph{dependency tree} is a tuple \tuple{\W, w_r, R_\D, V_\C} where
$R_\D$ forms a tree over $\W$ rooted in $w_r$ and $V_\C: \W \mapsto \C$
maps each word to a word class.
\end{definition}

In a similar way, non-atomic properties of words, such as morphosyntactic
features, may be described. Our approach to feature structures is very
similar to the one proposed by \ct{Blackburn-SLT}.

\subsection{Order Domain Structure}

We have abandoned projectivity of the dependency tree to retain the
semantic motivation of dependencies. To formulate order restrictions, we
now introduce word order domains. The word order domain structure is a
hierarchy of word order domains, which in turn are sets of words. We link
the dependency tree to the domain structure and require projectivity not of
the dependency tree, but of the domain structure. The flexibility of the
linking allows to represent word order variation including discontinuities
in the domain structure alone, keeping the dependency tree constant.

More precisely, a \emph{word order domain} is a set of words, whose
cardinality may be restricted to at most one element, at least one element,
or -- by conjunction -- to exactly one element. Each word is associated
with a sequence of order domains, one of which must contain the word
itself, and each of these domains may require that its elements have
certain morphosyntactic features. Order domains are partially ordered
based on set inclusion: If an order domain $d$ contains word $w$ (which is
not associated with $d$), every word $w'$ contained in a domain $d'$
associated with $w$ is also contained in $d$; therefor, $d' \subset d$ for
each $d'$ associated with $w$. This partial ordering induces a tree on
order domains, which we call the \emph{order domain structure}.

\begin{figure}[t]
\centering
\epsfig{file=depstruc.eps,width=0.5\columnwidth}
\hfill 
\epsfig{file=domtree.eps,width=0.35\columnwidth}
\caption{Dependency Tree (left) and Order Domain Structure (right) for
	\os{Den Mann hat der Junge gesehen} 
	\label{fig:trees}}
\end{figure}

Again, take the example of German \os{Den Mann hat der Junge gesehen}. The
left of Fig.\ref{fig:trees} shows the word order domains by dashed circles.
The finite verb, \os{hat}, defines a sequence of domains, \tuple{d_1, d_2,
d_3}, which roughly correspond to the topological fields in the German main
clause. The nouns and the participle each define a single order domain. Set
inclusion gives rise to the domain structure on the right of
Fig.\ref{fig:trees}, where the individual words are attached by dashed
lines to their including domains.%
\footnote{ 
Note that in this case we have not a single rooted tree, but rather an
ordered sequence of trees (by virtue of ordering $d_1, d_2$, and $d_3$) as
domain structure. In general, we assume the sentence period to govern the
finite verb and to introduce a single domain for the complete sentence. 
}

Surface order is derived from an order domain structure by propagating
precedence relations from domains to their elements, i.e., \os{Mann}
precedes (any element of) $d_2$, \os{hat} follows (any element of) $d_1$,
etc. Order within a domain, e.g., of \os{hat} and $d_6$, or $d_5$ and
$d_6$, is based on precedence predicates. There are two different types,
one ordering a word w.r.t. any other element of the domain it is associated
with (e.g., \os{hat} w.r.t. $d_6$), and another ordering two modifiers,
referring to the dependency relations they occupy ($d_5$ and $d_6$,
referring to \texttt{subj} and \texttt{vpart}). A verb like \os{hat}
introduces two precedence predicates, requiring other words to follow
itself and the participle to follow subject and object, resp.:%
\footnote{
For details of the notation, please refer to \cite{nobi:coling98}.
}

\begin{center}
\os{hat} $\IMPL (<_{*} \AND\ \tuple{vpart} >_{\{subject, object\}})$
\end{center}

Informally, the first conjunct is satisfied by any domain in which no word
precedes \os{hat}, and the second conjunct is satisfied by any domain in
which no subject or object follows a participle. The domain structure in
Fig.\ref{fig:trees} satisfies these restrictions since nothing follows
the participle, and because \os{den Mann} is not an element of $d_2$, which
contains \os{hat}. This is an important interaction of order domains and
precedence predicates: Order domains define scopes for precedence
predicates. In this way, we take into account that dependency trees are
flatter than PS-based ones.%
\footnote{
Note that each phrasal level in PS-based trees defines a scope for linear
precedence rules, which only apply to sister nodes. 
}

Order domains easily extend to discontinuous dependencies. Consider the
non-projective tree in Fig.\ref{fig:trees}. Assuming that the finite verb
governs the participle, no projective dependency between the object \os{den
Mann} and the participle \os{gesehen} can be established. We allow
non-projectivity by loosening the linking between dependency tree and
domain structure: A modifier (e.g., \os{Mann}) may not only be inserted
into a domain associated with its direct head (\os{gesehen}), but also into
a domain of a transitive head (\os{hat}), which we will call the
\emph{positional head}.

The possibility of inserting a word into a domain of some transitive head
raises the questions of how to require continuity (as needed in most
cases), and how to limit the distance between the governor and the
modifier. Both questions are solved with reference to the dependency
relation. From a descriptive viewpoint, the \emph{syntactic construction}
is often cited to determine the possibility and scope of discontinuities
\cite{Bhatt1990,Matthews1981}. In PS-based accounts, the construction is
represented by phrasal categories, and extraction is limited by bounding
nodes (e.g., \ct{Haegeman1994,Becker1991}). In dependency-based
accounts, the construction is represented by the dependency relation, which
is typed or labelled to indicate constructional distinctions which are
configurationally defined in PSG. Given this correspondence, it is natural
to employ dependencies in the description of discontinuities as follows:
For each modifier, a set of dependency types is defined which may link the
direct head and the positional head of the modifier (\os{gesehen} and
\os{hat}, respectively). If this set is empty, both heads are identical and
a continuous attachment results. The impossibility of extraction from,
e.g., a finite verb phrase may follow from the fact that the dependency
embedding finite verbs, \texttt{propo}, does not appear on any path between
a direct and a positional head.

Formally, we define order domains and domain structures as follows.

\begin{definition}{Order Domain}
An \emph{order domain} is a continuous subset of \W, i.e., for any two
words contained in the order domain, all words in between are also
contained in the order domain. 
\end{definition}

\begin{definition}{Order Domain Structure}
An \emph{order domain structure} \M\ is a set of order domains which
satisfy the following restrictions: First, set inclusion defines a
hierarchy over \M:
\[
\forall m,m' \in M: 
        m \subseteq m' \OR m' \subseteq m \OR m \cap m' = \emptyset.
\]
Second, the top element of this hierarchy is equal to \W, i.e., \M\
contains all words.
\end{definition}

These definitions ensure that the domain structure defines a partial order
over the words, which can be extended to a total ordering by adding
precedence restrictions on elements within one domain. It is thus similar
to projective context-free trees, albeit without any categorial
information. This will be crucial later on.

\subsection{Dependency Structures}
	\label{ch:linking}

We still need to link the dependency tree to the dimensions of feature
structure and domain structure. This is achieved by the following
definition.

\begin{definition}{Dependency Structure}
A \emph{dependency structure} $T$ is a tuple \tuple{\W, w_r, R_\D, V_\C,
\P, R_\F, V_\A, V_\P, \M, V_\M} where \tuple{\W, w_r, R_\D, V_\C} is a
dependency tree, \tuple{\P, R_\F, V_\A} is a feature structure, and \M\ is
an order domain structure over \W. $V_\P: \W \mapsto \P$ maps each word to
a point in the feature structure, $V_\M: \W \mapsto \M^n$ maps each word to
a sequence of order domains.
\end{definition}

Besides other restrictions not relevant here, we require of a dependency
structure four more conditions: (1) Each word $w \in \W$ is contained in
exactly one of the domains from $V_\M(w)$, (2) all domains in $V_\M(w)$ are
pairwise disjoint, (3) each word (except $w_r$) is contained in at least
two domains, one of which is associated with a (transitive) head, and (4)
the (partial) ordering of domains (as described by $V_\M$) is consistent
with the precedence of the words contained in the domains.

\section{Comparison with LFG
	\label{ch:compare}}

We now turn to comparing the \dachs\ approach to LFG. We have already noted
several shared basic assumptions in Section~\ref{ch:LFG}, so we will pick
out three differences here; Lexicalization, projections and their linking,
and word order.

\subsection{Lexicalization}

A recent tendency in linguistics is the attempt to move more and more
linguistic information into the lexicon, thereby eliminating rule systems.
Apart from linguistic reasons to do so (which may be debatable), there are
some very practical consequences of lexicalization. Large rule systems have
proven unwieldy over and over, resulting in unforeseen interactions and
questions of where to put (and, equally important, find) certain
descriptions. The development of lexica structured by inheritance makes it
possible to completely lexicalize a grammar without introducing redundancy.
Linguistic generalizations are expressed by class formation, but eventually
all grammatical information is located at individual words.

Here we see a clear difference between \dachs\ and LFG. \dachs\ is --
similar to many other DGs -- strictly lexicalized, i.e., there are no
rules, but only one combination operation which constructs a larger
dependency structure from two smaller ones. Although a lot of information
can be moved to lexical items, LFG retains a large set of rules for
c-structure. We think this is a disadvantage at least when it comes to
writing large grammars.

Lexicalization may also reduce the processing cost, as \cite{Schabes1988}
argue. This can be the case only if lexicalization itself does not
introduce ambiguities. Unfortunately, this is quite often the case because
the non-determinism implicit in rule selection has to be made explicit in
lexical entries. Due to this reason, L-TAG (and also some variants of CG)
suffers from an increase in lexical ambiguity of factor 10
\cite{Joshi+Srinivas1994}! In our view, this results from the combined
description of several information types (in this case, category,
dominance, and precedence) on one level, represented by the elementary tree.
\dachs\ does not suffer from this increas of lexical ambiguity, because the
levels each encode one type of information, and alternatives on one level
need not (although sometimes they must) be multiplied into other levels.

\subsection{Projection Architecture}

Besides the general similarity in using several linked dimensions or
projections, there are striking similarities in their informational
content. For example, the dependency tree quite closely corresponds to
LFG's f-structure: Both are unordered hierarchies representing
subcategorization whose relations (dependencies vs. grammatical functions)
are even similar. The feature structure which we did not discuss
corresponds to the m(orphological)-structure proposed by \cite{Butt1996}
for LFG, since both encode morphosyntactic information. Less similarity
must be noted for the conceptual structure of \dachs\ and s-structure of
LFG \cite{Halvorsen1995}, which have a slightly different motivation. As we
will discuss below, there are also some similarities between the domain
structure and the c-structure, in that both encode order restrictions.

The major difference between the \dachs\ and LFG architectures in this area
is the linking of levels and their status. \dachs\ identifies one level,
the dependency tree, as the fundamental one and maps other levels off it,
whereas LFG assumes all levels to be orthogonal and of equal importance.
Correspondingly, the linking relations are different: In \dachs, one might
draw the level dependencies as on the left of Fig.~\ref{fig:archs}, while
\cite{Kaplan1995} gives the diagram on the right as (one possible)
architecture of LFG (with the m-structure of \cite{Butt1996} added;
\cite{Halvorsen1995} maps s-structure directly off of c-structure).

\begin{figure}
\centering
\epsfig{file=dachs-dims.eps,width=0.4\textwidth}
\hfill
\epsfig{file=lfg-proj.eps,width=0.5\textwidth}
\caption{Projection Architecture in \dachs\ (left) and LFG (right)
	\label{fig:archs}}
\end{figure}

\subsection{Description of Word Order}
	\label{ch:c-struc}

The main point we want to make concerns c-structure. It usually is
described to represent more language-specific information than f-structure.
C-structure encodes two types of information; categories and precedence.
For languages such as German, which exhibit a quite free word order, it is
questionable whether confounding these two information types is
linguistically motivated and results in readable specifications.

We have already sketched in Section~\ref{ch:german} the problems resulting
from the combination of categorial and linear restrictions in one level. In
contrast, \dachs\ defines category-independent order domains which
explicitly recognizes topological fields. Restrictions on the cardinality
of the fields may be directly specified without reference to the field's
elements. A major achievement in our view is that word order variation is
not represented by rules (as in LFG) or by lexical ambiguity (as in L-TAG
\cite{Joshi+Srinivas1994} and some versions of CG \cite{Hepple1994}), but
rather by disjunctive descriptions of a separate dimension. This not only
allows a concise description of the linguistic notions behind precedence
and topological fields (because they are stated in a language specially
suited for this specification), but also eliminates alternatives for
parsing, reducing the search space. In a way, this recapitulates the
ambiguity-reducing effect of feature annotations for order restrictions.







\section{Conclusion
	\label{ch:conclu}}

This paper has explored an architecture based on the dependency relation
which exhibits a number of similarities to LFG. As in LFG, we view
restrictions on morphosyntactic features, word order, and conceptual
interpretation as largely independent, similar to the modularity assumption
for LFG projections. In contrast to LFG, we see them as consequences (or,
for analytical purposes such as parsing, as indicators) of an abstract
syntactic primitive. Consequently, this abstract syntactic representation
is given special status, which shows up in, e.g., the star-shaped
projections originating on this fundamental level, as opposed to the more
linear projection architecture of LFG.

We have argued that the information content on each projection or dimension
should be restricted to one type, and that confounding categorial and
precedence information in c-structure has undesirable consequences. These
consequences materialize in complex rule systems (where large sets of
categories must be enumerated in certain positions) or -- in lexicalized
theories -- in lexical ambiguities (representing word order variation). We
think that the projection idea allows a better separation of alternatives
than in LFG and have defined -- as a separate projection -- word order domains,
which have no categorial implications. One feature of word order domains is
that they factor ordering alternatives from the syntactic tree, much like
feature annotations do for morphological alternatives. The traditional
description in terms of semantically motivated dependencies and topological
fields has been backed up by a state-of-the-art formal framework, which is
based on modal logic.

In the light of this work, it seems valid to reconsider the dichotomy
between PS-based and dependency-based approaches to language. Very
generally, it could be argued that PSG -- besides the non-lexical
categories -- requires a notion of valency and additional machinery to
cover order variation, whereas DG is already based on valency and only
requires an ordering component such as the one sketched in this
article. Perhaps one could eliminate \os{the nonobservable linguistic
construct that enjoys the widest acceptance} \cite[:9, referring to
nonlexical categories]{PS-II} by investing more work in DG.

\smallbibliography{\small}
\bibliographystyle{nobibib}

\end{document}